\documentclass{article}

 \PassOptionsToPackage{numbers, compress}{natbib}
\usepackage[final]{nips_2017}

\usepackage[utf8]{inputenc} 
\usepackage[T1]{fontenc}    
\usepackage{hyperref}       
\usepackage{url}            
\usepackage{booktabs}       
\usepackage{amsfonts}       
\usepackage{nicefrac}       
\usepackage{microtype}      
\usepackage{subfigure}
\usepackage{graphicx} 
\usepackage{floatrow}

\usepackage{hyperref}

\usepackage{amsmath}
\usepackage{amssymb}
\usepackage{amsthm,dsfont}

\newcounter{theorem}
\newtheorem{lemma}[theorem]{Lemma}

\DeclareMathOperator*{\argmin}{arg\,min}

\title{A comparative study of counterfactual estimators}

\author{
  Thomas Nedelec \\
  Criteo Research\\
  ENS Paris Saclay\\
  \And
   Nicolas Le Roux \\
  Criteo Research\\
\\
  \And
  Vianney Perchet\\
  ENS Paris Saclay\\
  Criteo Research\\
}
\begin{document}
\maketitle

\begin{abstract}
We provide a comparative study of several widely used off-policy estimators (\emph{Empirical Average}, \emph{Basic Importance Sampling} and \emph{Normalized Importance Sampling}), detailing the different regimes where they are individually suboptimal. We then exhibit properties optimal estimators should possess. In the case where examples have been gathered using multiple policies, we show that fused estimators dominate basic ones but can still be improved.

\end{abstract}

\section{Introduction}
The Reinforcement Learning (RL) theory gathers approaches that enable autonomous agents to learn how to evolve in an unknown environment through trial-and-error feedbacks. These algorithms optimize the behavior of an agent according to rewards received through past interactions with the world (\cite{bertsekas1995neuro, sutton1998introduction}). Recently, RL has had success with implementing agents which learned to control a remote helicopter or play several Atari games without any prior knowledge on the environment. Very recently, it was a key block in AlphaGo, the first algorithm able to beat a human world master at Go.

To interact with its environment, an autonomous agent follows a policy  dictating the action to take, accordingly to some prescribed distribution. The expected reward of a policy $\pi$ is then defined as
 \begin{align*}
J(\pi) = \int_{\mathcal{T}} \pi(\tau)\bar{r}(\tau) \; d\tau.
\end{align*}
where $\mathcal{T}$ is the set of all actions $\tau$,  $\bar{r}(\tau)$ is the expected reward associated with  $\tau$ and $\pi$ is the distribution prescribed by the policy. A possible way to estimate $J(\pi)$ is to sample from $\pi$ and collect, for each rollout  $i \in \{1,\ldots N\}$, the chosen action $\tau_{i}$, and the reward $r_{i}$ whose conditional expectation is $\bar{r}(\tau_i)$. Then, noting $D^{N} = \{(\tau_{1}, r_{1}), ..., (\tau_{N}, r_{N})\}$ the sequence of actions and collected rewards, we may use the classical Monte-Carlo estimator: $\widehat{J}_{MC}(D^{N}) = \frac{1}{N} \sum_{i}^{N} r_{i}$.

However, in many settings such as robotics or industrial applications, it can be crucial to estimate the expected reward of a policy $\pi_{\textrm{test}}$ without sampling from it, as it may be too expensive (in time or money). Thus, the estimation   of a new policy has to be based on data gathered with a previous policy $\pi$, usually called the behavior policy in the RL community.

Offline methods were developed to use data from the behavior policy to evaluate the expected reward of a test policy (also called the target policy). This setting is known as \emph{off-policy  evaluation} (OPE) or \emph{counterfactual reasoning} ~\cite{bottou2013counterfactual}. Over the years, many estimators of the performances of a test policy have been developed, amongst which \emph{Basic Importance Sampling} (BIS,~\cite{hammersley5monte}), \emph{Normalized Importance Sampling} (NIS,~\cite{powell1966weighted}), \emph{Empirical Average} (EA,~\cite{hirano2003efficient}) and \emph{Capped Importance Sampling} (CIS,~\cite{bottou2013counterfactual}).

All these estimators achieve a different tradeoff between bias and variance and the standard way to compare them is through the use of the Mean Square Error (MSE), as mentioned by~\cite{thomas2016data}. However, when faced with a particular setup, there are no guidelines to choose a good estimator and one is often left with the task of trying them all.~\cite{li2014minimax} provided a first comparative study of \emph{basic importance sampling} with the \emph{empirical average} estimator but this was not extended to other popular estimators.

We study in section~\ref{sec:single} the differences between  BIS, NIS and EA with a single policy. In particular, we show that NIS may be seen as an interpolation between BIS, which we prove to be optimal when the rewards have high variance, and EA, which we prove to be optimal when the rewards are deterministic. We also make explicit desirable properties an estimator should have to achieve low MSE.

Even though these estimators were designed assuming all the examples were collected using a single policy $\pi$, they can be extended to the case where each sample $i$ has been collected using a different policy $\pi_i$ as shown in \cite{agarwal2017effective}.  Further, in section~\ref{sec:multiple}, we prove that, when examples have been collected using multiple policies, FIS dominates BIS. The proof is slightly different from the one provided in \cite{agarwal2017effective}. We then show that FIS is the optimal unbiased estimator when the variance of the rewards is very large but that, in the low variance regime, better performing unbiased estimators exist.

Let us now introduce some notations. First of all, for simplicity we will assume that the set of actions $\mathcal{T}$ is finite, even though our results (apart from those concerning (EA)) extend to the infinite case. Every time we select action $\tau$, we observe a random reward $r(\tau)$ with (unknown) expectation $\bar{r}(\tau)$ and variance $V_r(\tau)$. The objective is to estimate the expected reward of the target policy $\pi_{\textrm{test}}$:
\begin{align}
\label{eq:J_test}
J(\pi_{\textrm{test}}) = \sum_{\tau \in \mathcal{T}} \pi_{\textrm{test}}(\tau)\bar{r}(\tau).
\end{align}
We consider we have collected $\{(\tau_{i},r_{i})\}_{i \in [N]}$ where the sequence of actions $s=\{\tau_{i}\}_{i \in [N]}$, that we call  the sampled path, was generated by following the behavior policy $\pi$.

\section{Examples collected with a single policy}
\label{sec:single}
In this section, we assume that all the examples were collected using the same behavior policy $\pi$. Further, we will assume most of the time that all actions have been selected at least once, so that EA is well defined.

We recall the formula for the estimators we consider:
\begin{align*}
\widehat{J}_{\textrm{BIS}}(\pi_{\textrm{test}}, \pi, D^N) &= \frac{1}{N} \sum_{i=1}^{N} \frac{\pi_{\textrm{test}}(\tau_{i})}{\pi(\tau_{i})}r_{i}\\
\widehat{J}_{\textrm{NIS}}(\pi_{\textrm{test}}, \pi, D^{N}) &= \frac{1}{\sum_{i=1}^{N} \frac{\pi_{\textrm{test}}(\tau_{i})}{\pi(\tau_{i})}} \sum_{i=1}^{N} \frac{\pi_{\textrm{test}}(\tau_{i})}{\pi(\tau_{i})}r_{i}\\
\widehat{J}_{\textrm{EA}}(\pi_{\textrm{test}}, \pi, D^{N}) &= \sum_{\tau \in \mathcal{T}} \pi_{\textrm{test}}(\tau)\hat{r}(\tau) \; ,
\end{align*}
where, for EA, $\hat{r}(\tau)$ is the empirical average\footnote{In most implementations, when action $\tau$ has never been sampled, $\hat{r}(\tau)$ is set to 0.} of the rewards of action $\tau$. To exhibit the difference between these estimators, we rewrite them under the form
\begin{align}
\label{eq:JZ_with_omega}
\widehat{J}_Z &= \sum_{\tau \in \mathcal{T}} \omega_Z(\tau, s)\pi_{\textrm{test}}(\tau) \hat{r}(\tau) \; , \ \text{ with }
\end{align}

$$\omega_{\textrm{BIS}}(\tau, s) = \frac{k(\tau, s)}{N \pi(\tau)}\ , \ 
\omega_{\textrm{NIS}}(\tau, s) = \frac{k(\tau, s)}{\pi(\tau)\sum_{j=1}^{N} \frac{\pi_{\textrm{test}}(\tau_j)}{\pi(\tau_j)}}\ \text{ and }
\omega_{\textrm{EA}}(\tau, s) = 1 \; ,$$
and where $\hat{r}(\tau)$ is the empirical average of action $\tau$ and $k(\tau, s) = \sum_{\tau_i \in s} 1_{\tau_i = \tau}$
is the number of times action $\tau$ has been sampled in  $s$.
We shall compare these weights $\omega$ to the theoretical weights the estimator minimizing the MSE would have, which will now be computed.

\subsection{Theoretical optimal weights}
Since the samples are exchangeable, the dependency of the optimal weights $\omega^*(\tau, s)$ on $s$ is only limited to  $\big(k(\tau,s)\big)_{\tau \in \mathcal{T}}:=k(s) \in \mathds{N}^{\mathcal{T}}$, the list of all counts for a given path $s$. We also denote $\mathcal{K}$ the set of all possible such $k(s)$. Thus, we can rewrite the MSE as
$$MSE(\widehat{J}) = \mathbb{E}_{R,s}\bigg[ \bigg(\widehat{J} - J(\pi_{\textrm{test}})\bigg)^2\bigg]
= \sum_{\kappa \in \mathcal{K}} \sum_{s: k(s) = \kappa} \pi(s) \mathbb{E}_{R}\bigg[ \bigg(\widehat{J} - J(\pi_{\textrm{test}})\bigg)^2\bigg]
$$
As  optimal weights only depend  on $k(s)$, we can optimize independently each $\kappa \in \mathcal{K}$. Moreover, for every $\kappa \in \mathcal{K}$, $\mathbb{E}_{R}\bigg[ \bigg(\widehat{J} - J(\pi_{\textrm{test}})\bigg)^2\bigg]$ is constant for all paths $s$ such that $k(s) = \kappa$.
The problem therefore becomes
\begin{eqnarray*}
\omega^*(k_{1K}) &=& \argmin_{\omega} \mathbb{E}_{R}\bigg[ \bigg(\widehat{J} - J(\pi_{\textrm{test}})\bigg)^2\bigg]\\&=& \argmin_{\omega} \mathbb{E}_{R}\bigg[ \bigg(\sum_{\tau \in \mathcal{T}} \omega(\tau, s)\pi_{\textrm{test}}(\tau) \hat{r}(\tau) - \sum_{\tau \in \mathcal{T}}\pi_{\textrm{test}}(\tau) \bar{r}(\tau)\bigg)^2\bigg]\\
&=& \argmin_{\omega} \left(\sum_{\tau} (\omega(\tau, s) - 1) \pi_{\textrm{test}}(\tau) \bar{r}_{\tau}\right)^{2} + \sum_{\tau} \omega^{2}(\tau, s) \pi_{\textrm{test}}^{2}(\tau) \frac{V_{r}(\tau)}{k_{\tau}} \; .
\end{eqnarray*}
These optimal weights can be computed analytically (the calculation is provided in the appendix) and are equal to
\begin{align*}
\omega^*(\tau, k(s)) = \frac{k_{\tau}\bar{r}(\tau)}{\pi_{\textrm{test}}(\tau)V_{r}(\tau)} \frac{\sum_{\tau'} \pi_{\textrm{test}}(\tau')\bar{r}(\tau')}{1 + \sum_{\tau'} \frac{k_{\tau'}\bar{r}(\tau')^2}{V_{r}(\tau')}} ,
\end{align*}
where we used the notation $k_\tau = k(\tau,s)$ to simplify notations. We emphasize that these weights are only theoretical since $\bar{r}(\tau)$ and $V_{r}(\tau)$ are unknown.

In the case of a single action, this simplifies to
\begin{align}
\label{eq:optimal_single}
\omega^*(\tau, k_\tau) = \frac{\bar{r}(\tau)^{2}}{\bar{r}(\tau)^{2} + \frac{V_{r}(\tau)}{k_{\tau}}} \; .
\end{align}

%
%

Moreover, in that case, an unbiased estimator requires $E_{k_\tau}[\omega_Z(\tau, k_\tau)] = 1$. We see that, when the variance $V_{r}(\tau) / k_\tau$ is large, the optimal weight trades off variance for bias.

When $V_r(\tau) = 0$ we recover that the weights should be constant equal to one. Indeed in this case, the term appearing in the MSE is the bias term and the weights that are setting the bias to zero are constant and equal to one. These weights correspond to the \emph{empirical average} weights.

When $V_r(\tau)/\bar{r}^2(\tau)$ is high, we find that the optimal weight should depend on $k_{\tau}$ and $\frac{\bar{r}^2(\tau)}{V_r(\tau)}$. Intuitively, the bias should be higher when the variance of $\hat{r}(\tau)$ is very high. This variance depends both on the intrinsic variance of the reward and the number of times the action was taken. That is why the optimal weights depend on $k_{\tau}$ and $\bar{r}(\tau)$.

\subsection{Suboptimality of traditional counterfactual estimators}
\label{sec:suboptimality}
We now explore in which settings BIS, EA, NIS are suboptimal. To that extent, it is beneficial to realise where the variance of these estimators comes from.
There are two sources of variance in a counterfactual estimator. The first one comes from the variance of the rewards $V_{r}(\tau)$ and the second one comes from the variance of the path induced by the behavior policy. These two components can be made explicit by computing the variance of any estimator using the law of total variance (detailed in the appendix):
\begin{align}
\mathbb{V}(\widehat{J}_Z)&= V_{\textrm{int}}(\widehat{J}_Z) + V_{\textrm{path}}(\widehat{J}_Z) \text{ with }
\end{align}
$$
V_{\textrm{int}}(\widehat{J}_Z) = \mathbb{E}_{s \sim \pi}\left[\sum_{\tau} \omega_{\textrm{Z}}^{2}(\tau, s)\pi_{\textrm{test}}^2(\tau)\frac{V_{r}(\tau)}{k_{\tau}}\right] \  \text{and}\ 
V_{\textrm{path}}(\widehat{J}_Z) = \mathbb{V}_{s \sim \pi}\left[\sum_{\tau} \omega_{\textrm{Z}}(\tau, s)\pi_{\textrm{test}}(\tau) \bar{r}(\tau)\right] \; .
$$

$V_{\textrm{path}}$ is equal to $0$ when the weights $\omega$ are independent of the path $s$, as is the case with EA. $V_{\textrm{int}}$ is small when weights are small for the actions whose empirical average reward has high variance. However, since the latter depends on the number of times the action has been drawn, and thus on $s$, each estimator achieves a different tradeoff between $V_{\textrm{path}}$ and $V_{\textrm{int}}$.

\subsubsection{Basic Importance Sampling}
We recall that $\displaystyle \omega_{\textrm{BIS}}(\tau, s) = k_{\tau}/\pi(\tau)N$ and these weights are  linear in $k_\tau$. This linear relationship is the one found in the optimal weights of Eq.~\ref{eq:optimal_single} when the variance $V_{r}(\tau)$ is much larger than $\bar{r}^2(\tau)$. So, in the case of high variance $V_{r}(\tau)$, we expect BIS to be close to optimal. In particular, BIS has the desirable property that action sampled many times, and whose average reward is well estimated, have a higher weight in the final estimator than actions with poorly estimated average reward. In the low variance regime, $V_{\textrm{path}}$ dominates $V_{\textrm{int}}$. Since this estimator does not take the sampled path into account to reduce $V_{\textrm{path}}$, \textbf{BIS  is suboptimal in that low variance regime}. We exhibit experiments in Figure~\ref{fig:mse_for_varying_p} where we can observe the suboptimality of BIS when $V_{r}(\tau)$ is small.




\subsubsection{Empirical Average}
We  recall that the weights of the EA estimator are $\displaystyle \omega_{\textrm{EA}}(\tau, s) = 1$.
They are equal to the optimal weights of Eq.~\ref{eq:optimal_single} when $V_{r}(\tau)=0$. Indeed, if  rewards are deterministic, $V_{\textrm{path}}$ dominates $V_{\textrm{int}}$. Since the constant weights of EA induce $V_{\textrm{path}}=0$, the EA estimator is optimal in that regime, provided each action was sampled at least once. If, however, the variance of the rewards $V_{r}(\tau)$ is large, then $V_{\textrm{int}}$ dominates $V_{\textrm{path}}$. Since it  focuses on setting $V_{\textrm{path}}$ to 0 at the expense of a larger $V_{\textrm{int}}$, \textbf{EA  is suboptimal in that high variance regime}. Instead, we would like to downweight the actions which have been rarely sampled and upweight those which have been sampled often. Figure~\ref{fig:mse_for_varying_p}, we show experiments that prove this suboptimal behavior of EA when $V_r(\tau)$ is large

\subsubsection{Normalized Importance Sampling}
\label{sec:NIS}
We now focus our attention on NIS and show that this estimator may be seen as an interpolation between BIS and EA. First, we recall that the weights of the NIS estimator are
\begin{eqnarray*}
\omega_{\textrm{NIS}}(\tau, s) &= k_\tau\Big/\bigg(\pi(\tau)\sum_{j=1}^{N} \frac{\pi_{\textrm{test}}(\tau_j)}{\pi(\tau_j)}\bigg) \; ,
\end{eqnarray*}
which can be rewritten
%
%
%
%
\begin{align*}
\omega_{\textrm{NIS}}(\tau, s) &= \frac{k_{\tau}}{\pi(\tau)N}\Big/\bigg(\sum_{\tau \in \mathcal{T}}\pi_{\textrm{test}}(\tau)  \frac{k_{\tau}}{\pi(\tau)N}\bigg)
 = 1\Big/\bigg[\pi_{\textrm{test}}(\tau)\Big(1 + \frac{\sum_{\tau' \ne \tau}  \pi_{\textrm{test}}(\tau') \frac{k_{\tau'}}{\pi(\tau')N}}{\frac{\pi_{\textrm{test}}(\tau)k_{\tau}}{\pi(\tau)N}}\Big)\bigg].
\end{align*}

To simplify the analysis, we make the assumption that $\sum_{\tau' \ne \tau}  \pi_{\textrm{test}}(\tau') \frac{k_{\tau'}}{\pi(\tau')N} \approx 1 -  \pi_{\textrm{test}}(\tau)$ which is true in expectation if the behavior policy does not depend on the quality of the arms.
Under this assumption,
\begin{align}
\omega_{\textrm{NIS}}(\tau, s) &= k_\tau \Big/\bigg(k_\tau \pi_{\textrm{test}}(\tau) + (1 - \pi_{\textrm{test}}(\tau))\pi(\tau)N\bigg) \; .
\end{align}
$\omega_{\textrm{NIS}}(\tau, s)$ is a weighted harmonic average between $\omega_{\textrm{BIS}}(\tau, s)$ and $\omega_{\textrm{EA}}(\tau, s)$. Interestingly, while we would ideally like to interpolate between BIS and EA based on the variance of the rewards, the weight of NIS depends on $\pi_\textrm{test}$ instead.

In Table~\ref{table:NIS}, we compute the value of $\omega_{\textrm{NIS}}$ for different value of $\pi_{\textrm{test}}(\tau)$ based on this approximation.
\begin{table}[h!]
\centering
\caption{Approximation of the normalized weights}
\label{table:NIS}
\begin{tabular}{|c||c|c|c|}
\hline
$\pi_{\textrm{test}}(\tau)$&  $ \varepsilon$     & $0.5 $ &  $1 - \varepsilon $  \\ \hline
$\omega_{\textrm{NIS}}(\tau)$&$\frac{k_\tau}{\pi(\tau)N}$   & $\frac{2}{1 + \frac{\pi(\tau)N}{k_\tau}}$           &    1            \\ \hline
\end{tabular}
\end{table}

We show in Figure~\ref{fig:mse_for_varying_p} this interpolation by plotting the MSE of the three estimators as a function of the variance of the rewards. It makes it clear that
\begin{itemize}
\item[i)] the \emph{empirical average} estimator is optimal and  \emph{normalized important sampling} is better than \emph{basic importance sampling} when $V_{int}$ is low,\vspace{-0.1cm} 
 \item[ii)] \emph{basic importance sampling} is better than \emph{empirical average} when $V_{int}$ is high\vspace{-0.1cm} 
  \item[iii)] NIS achieves a tradeoff between \emph{empirical average} and \emph{normalized importance sampling}.
\end{itemize} We now present one experiment to show these different properties.


\subsection{Experiment with one behavior policy}
\label{sec:experiments}
We consider an environment with $K=20$ actions where each action yields rewards following a scaled Bernoulli distribution, i.e., $r(\tau) =         \frac{Z(\tau)}{\sqrt{p}}$ with probability $p$ and 0 otherwise, with $p \in [0,1]$. Indexing the actions from $1$ to $K$, we consider a symmetric reward defined as $Z(i) = i/K$ for $i=1:K/2$  and $Z(i) = Z(K - i)$ for $i=K/2:K$. Since $\bar{r}^2(\tau) = pZ^2(\tau)$ and $V_{r}(\tau) = (1-p)Z^2(\tau)$, varying $p$ from $0$ to $1$ changes the ratio $\bar{r}^2(\tau) / (\bar{r}^2(\tau) + V_{r}(\tau))$.

For the sampling policy, we consider $\pi(i) = \frac{2i}{K(K+1)}$. Thus, symmetric actions have the same   $V_r(\tau)$ but the action whose index is superior to $K/2$ is sampled more times than the action whose index is inferior to $K/2$. $\pi_{\textrm{test}}$ is a peaked distribution, choosing two actions with equal probability $0.475$ ($K=20$ and $K=10$) and the remaining actions with equal probability $\frac{0.05}{K-2}$. 

\begin{figure}
\floatbox[{\capbeside\thisfloatsetup{capbesideposition={left,center},capbesidewidth=.5\textwidth}}]{figure}[\FBwidth]
{\caption{MSE of \emph{BIS}, \emph{NBIS} and \emph{EA} as a function of $\displaystyle \frac{\bar{r}^2(\tau)}{\bar{r}^2(\tau) + V_{r}(\tau)}$. NIS achieves a tradeoff between \emph{empirical average} and \emph{normalized importance sampling}.}\label{fig:mse_for_varying_p}}
{\includegraphics[width=.45\textwidth]{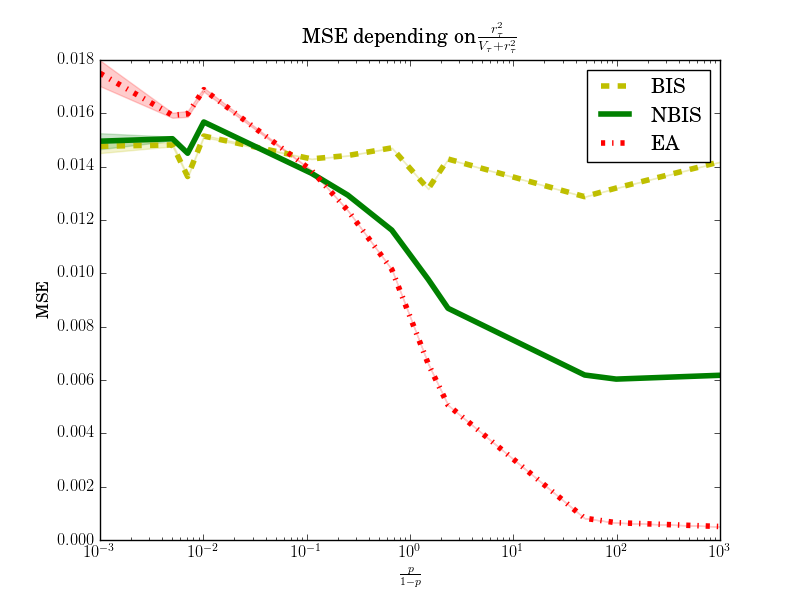}}
\end{figure}

Fig.~\ref{fig:mse_for_varying_p} shows the MSE of the estimators as a function of $p$. When rewards are almost deterministic (right part of each plot), we see the optimality of EA and the strong dependance of BIS on $V_{\textrm{path}}$. The gap between the MSE of EA and the MSE of BIS is in the right part of the plots corresponds to $V_{\textrm{path}}$. When $V_r(\tau)/\bar{r}_{\tau}^2$ is high, BIS and NIS achieve a lower MSE than EA since the weights of \emph{empirical average} do not depend on $k_{\tau}$ and suffer from a high $V_{\textrm{int}}$.

\section{Examples collected with multiple policies}
\label{sec:multiple}
We  extend our analysis to the case where different policies have been used to collect examples. Formally, we  consider a family of behavior policies $\{\pi_{i}\}_{i \in [1, N]}$ such that action $\tau_{i}$ was sampled according to $\pi_{i}$. In the same spirit of \cite{agarwal2017effective}, we show that, in this context, BIS is dominated by another estimator called \emph{Fused Importance Sampling} (FIS,~\cite{peshkin2002learning}). We then study how both of these estimators are suboptimal. Additionally, we provide a new unbiased estimator which theoretically outperforms the FIS. Finally, we detail why, in some cases with several policies  implemented, one must be careful when using the EA estimator.

\subsection{BIS and FIS in the context of multiple policies}
With multiple policies,  importance sampling techniques can  be used by considering the importance weights corresponding to the policy used to collect data. Corresponding estimators may be written
\begin{eqnarray*}
\label{eq:J_with_alpha_multiple}
\widehat{J}_Z(\pi_{\textrm{test}}, \{\pi_i\}_, D^N) &= \frac{1}{N}\sum_{i=1}^N \sum_\tau \alpha_Z^i(\tau, s)\frac{\pi_{\textrm{test}}(\tau)}{\pi_i(\tau)} r_i 1_{\tau_i = \tau} \; .
\end{eqnarray*}
where $\alpha_{\textrm{BIS}}^i(\tau, s)= \alpha_{\textrm{BIS}}^i(\tau) = 1$. To distinguish  from Eq.~\ref{eq:JZ_with_omega}, we denote  weights by $\alpha_Z$ instead of $\omega_Z$.

FIS uses another formulation for the weights, namely $\alpha_{\textrm{FIS}}^i(\tau, s) = \pi_i(\tau)/\frac{1}{N} \sum_{j=1}^{N}{\pi_j(\tau)}$.
FIS is usually preferred to BIS but, until now, there was no theoretical justification for this choice (the result is briefly mentioned in \cite{shelton2001importance}). Lemma~\ref{lemma:FIS_dominates}, whose proof is delayed to appendix, proves that FIS dominates BIS.

\begin{lemma}[FIS dominates BIS]
\label{lemma:FIS_dominates}
Assume  $N$ policies $\pi_1, \ldots, \pi_N$ sampled each an action $\tau_i$ and received a random reward $r_i(\tau_i)$. To assess the average reward obtained using a test policy $\pi_{\textrm{test}}$, define the two estimators, \emph{Basic Importance Sampling} and \emph{Fused Importance Sampling} by:
$$\widehat{J}_{\textrm{BIS}} = \frac{1}{N} \sum_{i=1}^N \int_{\tau} \frac{\pi_{\textrm{test}}(\tau)}{\pi_i(\tau)}r_i(\tau) 1_{\tau_i = \tau} \; d\tau  \quad \text{and} \quad
\widehat{J}_{\textrm{FIS}} = \sum_{i=1}^N \int_{\tau} \frac{\pi_{\textrm{test}}(\tau)}{\sum_j \pi_j(\tau)}r_i(\tau) 1_{\tau_i = \tau} \; d\tau \; .
$$Then we have $\mathbb{V}(\widehat{J}_{\textrm{FIS}}) \leq \mathbb{V}(\widehat{J}_{\textrm{BIS}})$.
Further, since both estimators are unbiased, $\widehat{J}_{\textrm{FIS}}$ dominates $\widehat{J}_{\textrm{BIS}}$.
\end{lemma}
The intuition is the following. Consider a fixed action $\tau$ and assume that this action has been sampled several times, at stage $n_1, n_2, \ldots, n_k$. The overall  weight put by BIS is of the order of $$\frac{1}{N}\Big(\frac{\pi_{\textrm{test}}(\tau)}{\pi_{n_1}(\tau)} + \frac{\pi_{\textrm{test}}(\tau)}{\pi_{n_2}(\tau)} +\ldots  \frac{\pi_{\textrm{test}}(\tau)}{\pi_{n_k}(\tau)} \Big)$$ while the weight put by FIS would be of the order of $k\pi_{\textrm{test}}(\tau)/\sum_{j=1}^N \pi_j(\tau)$. 
If  one of the $\pi_{n_i}$ is really small (but not the other), then the weight put on BIS can be  huge while the one of FIS would remain reasonably small. Looking at the details of the proof, the key argument when comparing the variances is how different means (Cesaro vs harmonic means) compare with each other.


\subsection{The optimal unbiased estimator}
Even though FIS dominates BIS, it is not optimal and we provide a new estimator with a lower MSE.

\begin{lemma}[Optimal unbiased estimator]
Consider the family of estimators of the form:
\begin{eqnarray*}
\widehat{J}_Z(\pi_{\textrm{test}}, \{\pi_i\}, D^N) &= \sum_{i=1}^N \sum_{\tau \in \mathcal{T}} \alpha_Z^i(\tau )\frac{\pi_{\textrm{test}}(\tau)}{\pi_i(\tau)} r_i 1_{\tau_i = \tau}
\end{eqnarray*}
where for all $ \tau\in\mathcal{T}$,  it holds that $\sum_{i=1}^{N} \alpha_Z^i(\tau) = 1$.
Amongst this family, the weights of the estimator that minimizes the MSE can be written as:
\begin{eqnarray*}
\alpha_{opt}^i(\tau)= \frac{\pi_{i}(\tau)}{\bar{r}^2(\tau)(1-\pi_{i}(\tau)) + V_{r}(\tau)}/\sum_{j=1}^{N} \frac{\pi_{j}(\tau)}{\bar{r}^2(\tau)(1-\pi_{j}(\tau)) + V_r(\tau)}
\end{eqnarray*}
with $\bar{r}(\tau)$ and $V_r(\tau)$ as defined in the introduction.
\end{lemma}

\begin{proof}
As estimators must be unbiased, we only need to minimize the variance of $\widehat{J}_Z(\pi_{\textrm{test}}, \{\pi_i\}, D^N)$ with respect to $\{\alpha_Z^i(\tau)\}_{i \in [N], \tau }$. Since we are free to choose the weights for each action independently, we focus on minimizing the MSE computed on one action $\tau$. For a given action $\tau$, the estimator is the sum of N independent random variables. Focusing on the variance for one sample, we have
\begin{eqnarray*}
\mathbb{V}\bigg[\widehat{J}_{i}(\tau_{i}, r(\tau))\bigg] = \mathbb{V}\left[\frac{\alpha^i(\tau)\pi_{\textrm{test}}(\tau)}{\pi_{i}(\tau)}r(\tau)\textbf{1}(\tau = \tau_{i})\right] \; .
\end{eqnarray*}
We can use the law of total variance and compute:
\begin{eqnarray*}
\mathbb{V}\bigg[\widehat{J}_i(\textbf{1}(\tau_i = \tau), r(\tau))\bigg]
&=& \mathbb{V}_s\left[\mathbb{E}_R\left[\frac{\alpha^i(\tau)\pi_{\textrm{test}}(\tau)}{\pi_{i}(\tau)}r(\tau)\textbf{1}(\tau = \tau_{i})\right]\right]\\
&& \qquad + \mathbb{E}_s\left[\mathbb{V}_R\left[\frac{\alpha^i(\tau)\pi_{\textrm{test}}(\tau)}{\pi_{i}(\tau)}r(\tau)\textbf{1}(\tau = \tau_{i})\right]\right]
\\&=&  (\bar{r}(\tau)\alpha^i(\tau))^{2}\pi_{\textrm{test}}^{2}(\tau) \left(\frac{1}{\pi_{i}(\tau)} - 1\right)+ \frac{(\alpha^i(\tau)\pi_{\textrm{test}}(\tau))^{2}}{\pi_{i}(\tau)}V_{r}(\tau) \; .
\end{eqnarray*}
Since the unbiasedness requires $\sum_{i=1}^{N} \alpha^i(\tau) = 1$, we compute the Lagrangian:
$$\mathcal{L}(\alpha^i(\tau) , \lambda) = (\bar{r}(\tau)\alpha^i(\tau))^{2}\pi_{\textrm{test}}^{2}(\tau) \left(\frac{1}{\pi_{i}(\tau)} - 1\right)
 \frac{(\alpha^i(\tau)\pi_{\textrm{test}}(\tau))^{2}}{\pi_{i}(\tau)}V_{r}(\tau)
+ \lambda(\sum_{i=1}^{N} \alpha^i(\tau) - 1)\, 
$$  that we optimize to  find $\alpha_{opt}^i(\tau)$.
\end{proof}
In the sequel, we call this estimator the \emph{Optimal Unbiased importance sampling} estimator (OUIS).
We first remark that there exists a trade-off between $\bar{r}^2(\tau)$ and $V_{r}(\tau)$. When $\frac{V_{r}(\tau)}{\bar{r}^2(\tau)}$ is small, the optimal weights depend on $\pi_{i}(\tau)/(1-\pi_{i}(\tau))$ which differ from the weights corresponding to the fused distribution, linear in $\pi_{i}(\tau)$. This is striking when  collecting policies are peaked. However, when $\frac{V_{r}(\tau)}{\bar{r}^2(\tau)}$ is high, weights matchs those of \textit{the fused distribution}.  We proved that the weights of the latter are the unbiased weights with no dependence on the sampled path that are minimizing $V_{\textrm{int}}$. We also remark that when the rewards are deterministic, the optimal weights are equal to:
\begin{align}
\omega_Z^i(\tau) = \frac{\pi_{i}(\tau)}{1-\pi_{i}(\tau)}/\sum_{j=1}^{N} \frac{\pi_{j}(\tau)}{1-\pi_{j}(\tau)}
\end{align}
Here, we do not need to compute  estimates of $\bar{r}(\tau)$ and $V_{r}(\tau)$ to be able to use the weights.




\subsection{Dependence of sampling policies on previous observed rewards}

There exists a strong difference between EA and importance sampling based estimators when the sampling policies were dependent from rewards previously observed by the system. It is the case for instance if $\pi_1$ is dependent from the rewards gathered by sampling $\pi_0$. This dependence can appear when the sampling policies are the intermediate steps of some policy learning algorithms.

We claim that in this case importance sampling based methods still lead to unbiased counterfactual estimators whereas the \emph{empirical average} estimator can be biased. 
Indeed, consider a simple setting with one action whose reward is a Bernoulli of parameter 1/2. The policy stops sampling this action as soon as it experiences a 0. By doing the calculation, we can show that the expected value of EA is $1 - \log(2)$ and its bias equals $\log(2) - 1/2$ (we provide the details in the appendix).

On the other hand, important sampling based methods are unbiased even when collecting policies are dependent: this is a standard result in the adversarial bandit literature (e.g. \cite{bubeck2012regret}).

\subsection{Experiments with multiple policies}
\subsubsection{Cartpole environment}
To compare  the performances with multiple policies, we first test them on the Cartpole\footnote{https://gym.openai.com/} environment. We consider stochastic linear policies where at each time step the cart moves right with probability $\sigma(x^{T}\theta)$ where $x$ is the state of the environment and $\theta$ the parameter of the model. To optimize the reward of the agent, we use the PoWER algorithm~\cite{kober2009policy} and consider  policies that were used to collect data in the optimisation process. To test our estimators, we estimated the expected reward of the final policy reached by the optimisation algorithm with the data collected by the 10 previous implemented policies. In each experiment, we use 300 rollouts (30 rollout per policy) to compute the estimators.

We use the per-decision version of each estimatoy \cite{precup2000eligibility} and we we compute its RMSE by running this process 400 times. We compute confidence intervals by bootstrapping and give the value of the 5th, 50th, and 95th percentiles. For the capped estimators, we use 10 as capping parameter. We also tested the \emph{Normalized fused importance sampling} estimator as defined by \cite{shelton2001policy}.

\begin{table}[h!]
\centering
\label{my-label2}
\caption{RMSE of the different estimators (confidence intervals computed with 400 runs)}
\begin{tabular}{|l|c|c|c|}
\hline
      & RMSE (5th) & RMSE (mean) & RMSE(95th) \\ \hline
BIS   &     122.32       &  236.73   &   321.40        \\ \hline
BCIS  &      84.65       & 87.30      & 90.24            \\ \hline
NBIS  &    39.34         &   43.22    &      47.52       \\ \hline
NBCIS &     31.02        &  32.77     &     34.63        \\ \hline
FIS   &    23.10         &   25.38    & 27.91             \\ \hline
OUIS  &     23.13        &   25.82    &    28.77         \\ \hline
NFIS  &   7.06          &    7.91    &   8.66           \\ \hline
\end{tabular}
\end{table}
Our \emph{optimal unbiased estimator} has similar performance to the \emph{Fused Importance Sampling} estimator as the probability of most of the rollouts (except path of size one and 2) is tiny and  $\pi_{i}(\tau)/(1-\pi_{i}(\tau)) \approx \pi_{i}(\tau)$. Thus, the \emph{fused} weights are very close to the \emph{optimal unbiased} weights .

\subsubsection{Blackjack environment}

We ran similar experiments in a blackjack environment. We used a policy iteration algorithm to maximize the reward of the agent. The algorithm is a Monte-Carlo policy iteration algorithm that plays epsilon-greedy according to the current Q function.   As in the Cartpole example, we select several policies that were considered in the optimisation process. We run the policy iteration algorithm 5000 times and consider 10 policies corresponding respectively to the time steps multiple of 500. The task is to compute the expected reward of the final policy based on 1000 rollouts (100  per policy).

If the policies considered are similar, i.e, if for all $i,j$ and all $\tau$, $\pi_i(\tau) \approx \pi_j(\tau)$, then
all the considered estimators are similar. To avoid this case, we play on the exploration rate in the policy iteration algorithm. An exploration rate that is decreasing slowly will lead to non-similar policies. We tested two different schemes to decrease the exploration rate, namely  $\epsilon_{t}^{1} = \frac{2}{2 + \log{t}}$ and  $\epsilon_{t}^{2} = 1- \frac{0.95t}{n_{iter}}$,
where $n_{iter}$ is the number of iterations of the policy iteration algorithm. The policies used in the estimator must be more different in both cases and we should observe a higher difference between the \textit{fused importance sampling estimator} and the \textit{optimal unbiased estimator}. We have not implemented the capped estimators since the weights are not very high.
The confidence intervals are computed by bootstrapping based on 2000 runs of the experiment. The results are gathered below.
\begin{table}[h!]
\begin{center}
\begin{tabular}{l|c|c|c|p{1cm}|c|c|c|}
\multicolumn{1}{c}{} &\multicolumn{3}{c}{MSE with $\epsilon^{1}_{t}$ (2000 runs) }&\multicolumn{1}{c}{}&\multicolumn{3}{c}{MSE with $\epsilon^{2}_{t}$ (2000 runs)}\\
\multicolumn{1}{c}{} & \multicolumn{1}{c}{ 5th} & \multicolumn{1}{c}{ mean} & \multicolumn{1}{c}{95th}& \multicolumn{1}{c}{}     & \multicolumn{1}{c}{ 5th} & \multicolumn{1}{c}{ mean} & \multicolumn{1}{c}{95th} \\\cline{2-4}\cline{6-8} 
BIS   &  0.1251          &  0.1255    &    0.1261 & & 0.0905 & 0.0910 & 0.0915 \\ \cline{2-4}\cline{6-8} 
FIS   &      0.1249       &   0.1254    &    0.1258    &   &      0.0870       &   0.0875    &    0.0880     \\\cline{2-4}\cline{6-8} 
OUIS  &     0.1250        &   0.1253    &  0.1258  &&        0.0855        &   0.0860    &  0.0865       \\ \cline{2-4}\cline{6-8} 
NBIS  &        0.0427     &  0.0438     &      0.0450 &&           0.0523     &  0.0535     &      0.0546     \\ \cline{2-4}\cline{6-8} 
NFIS  &      0.0347        &   0.0354     &   0.0363  &&         0.0429        &   0.0441     &   0.0453     \\ \cline{2-4}\cline{6-8} 
NOUIS & 0.0363 & 0.0374 &  0.0383 && 0.0458    & 0.0468  & 0.0479 \\ \cline{2-4}\cline{6-8} 
\end{tabular}
\end{center}
\label{table:blackjack_1}
\end{table}
%
%
For the first exploration parameter,  policies used to compute the estimator are  similar  and differenced between OUIS and FIS cannot be observed. With  sufficiently different policies, significant differences between the two estimators can bd observed and OUIS has a lower MSE as expected. We also tested a normalized estimator based on the weights of OUIS but this estimator has higher MSE than NFIS. Having better weights in the unbiased case is not a guarantee to build a better normalized estimator.

\section{Conclusion}

Our work provides some key elements for understanding in which cases the different usual counterfactual estimators are suboptimal and why we can see \emph{normalized importance sampling} as an interpolation between \emph{empirical average} and \emph{basic importance sampling}. 

We also focused on estimators that are using data gathered by multiple policies. We proved that \emph{fused importance sampling} dominates \emph{basic importance sampling} and then exhibited a new estimator that dominates FIS. This estimator is the optimal unbiased estimator. However, finding a better estimator that trades off bias for variance is still an open question in the case of multiple policies. 

These estimators represent a way to build data efficient off policy learning algorithms since they can reuse all data gathered in the learning process. One of our further direction of research would be to see how they reduce the number of examples that need to be sampled and if they can be improved to speed up the convergence of the  different learning algorithms. 
\newpage
\bibliography{literature}
\bibliographystyle{plain}
\newpage
\appendix
\section{Proof: Law of total variance for deriving the estimators variance}
We prove that when:
\begin{align*}
\widehat{J}_{\textrm{Z}}(D^{N}) = \sum_{\tau} \omega_{Z}(\tau,s)\pi_{\textrm{test}}(\tau)\hat{r}(\tau).
\end{align*}
the variance can be written as:
\begin{align*}
\mathbb{V}(\widehat{J}_Z(D^{N})) &= \mathbb{E}_{s \sim \pi}\bigg[\sum_{\tau} \omega_{\textrm{Z}}^{2}(\tau,s)\pi_{\textrm{test}}^2(\tau) \frac{V_{r}(\tau)}{k_\tau}\bigg] + \mathbb{V}_{s _\sim \pi}\bigg[\sum_{\tau} \omega_{\textrm{Z}}(\tau,s) \pi_{\textrm{test}}(\tau) \bar{r}(\tau)\bigg].
\end{align*}
with $s$ the sampled path.
 \begin{proof}
 Following the law of total variance,
 \begin{align*}
 \mathbb{V}(\widehat{J}_Z(D^{N})) &= \mathbb{E}_s\bigg[\mathbb{V}_R[\widehat{J}_Z(D^{N})|s]\bigg] + \mathbb{V}_s\bigg[\mathbb{E}_R[\widehat{J}_Z(D^{N})|s]\bigg]
 \end{align*}
 with:
 \begin{align*}
 \mathbb{E}\bigg[\mathbb{V}[\widehat{J}_Z(D^{N})|s]\bigg] &= \mathbb{E}_{s \sim \pi}\left[\mathbb{V}\left[\sum_{\tau}\omega_Z(\tau,s)\pi_{\textrm{test}}(\tau)\hat{r}(\tau)\bigg|s\right]\right] =  \mathbb{E}_{s \sim \pi}\left[\sum_{\tau} \mathbb{V}\left[\omega_Z(\tau,s)\pi_{\textrm{test}}(\tau)\hat{r}(\tau)\bigg|s\right]\right]
 \\&=  \mathbb{E}_{s \sim \pi}\left[\sum_{\tau} \omega_Z^2(\tau,s)\pi_{\textrm{test}}^2(\tau) \mathbb{V}\left[\hat{r}(\tau)|s\right]\right]
=  \mathbb{E}_{s \sim \pi}\left[\sum_{\tau} \omega_Z^2(\tau,s)\pi_{\textrm{test}}^2(\tau) \frac{V_{r}(\tau)}{k_{\tau}}\right].
 \end{align*}
 and
 \begin{align*}
 \mathbb{V}\bigg[\mathbb{E}[\hat{r}(D^{N})|s]\bigg] &= \mathbb{V}_{s \sim \pi}\left[\mathbb{E}\left[\sum_{\tau} \omega_Z(\tau,s)\pi_{\textrm{test}}(\tau) \hat{r}(\tau) \bigg|s\right]\right] = \mathbb{V}_{s _\sim \pi}\left[\sum_{\tau} \omega_Z(\tau,s) \pi_{\textrm{test}}(\tau) \bar{r}(\tau)\right]
\end{align*}
\end{proof}

\section{Proof: Optimal weights with one collecting policy}

We show how to compute the weights $\{\omega_{opt}(\tau,s)\}$ which minimizes:
\begin{align*}
MSE(\widehat{J}(D^{N})) &= \left(\sum_{\tau} (\omega(\tau,s) - 1) \pi_{\textrm{test}}(\tau) \bar{r}(\tau)\right)^{2} + \sum_{\tau} \omega^2(\tau,s) \pi_{\textrm{test}}^2(\tau) \frac{V_{r}(\tau)}{k_{\tau}}
\end{align*}
They are equal to:
\begin{align*}
\omega_{opt}(\tau,s) = \frac{k_{\tau}\bar{r}(\tau)}{V_{r}(\tau)\pi_{\textrm{test}}(\tau)} \frac{\sum_{\tau'} \pi_{\textrm{test}}(\tau')\bar{r}(\tau')}{1 + \sum_{\tau'} \frac{k_{\tau'}\bar{r}(\tau')^2}{V_{r}(\tau')}}
\end{align*}
\begin{proof}
The MSE is quadratic in $\omega(\tau,s)$:
\begin{align*}
\frac{\partial MSE(\widehat{J}(D^{N}))}{\partial\omega(\tau,s)} &= 2\pi_{\textrm{test}}(\tau)\bar{r}(\tau)\left(\sum_{\tau'} (\omega(\tau',s) - 1) \pi_{\textrm{test}}(\tau') \bar{r}(\tau')\right) + 2\omega(\tau,s)\pi_{\textrm{test}}^{2}(\tau)\frac{V_{r}(\tau)}{k_{\tau}}
\end{align*}
and
\begin{align*}
\omega_{opt}(\tau,s) = -\frac{k_{\tau}\bar{r}(\tau)}{V_{r}(\tau)\pi_{\textrm{test}}(\tau)} \sum_{\tau'}(\omega_{opt}(\tau',s) - 1)\pi_{\textrm{test}}(\tau')\bar{r}(\tau')
\end{align*}
We note $\lambda = \sum_{\tau'} \omega_{opt,s}(\tau')\pi_{\textrm{test}}(\tau')\bar{r}(\tau')$.
We have:
\begin{align*}
\omega_{opt}(\tau,s)\pi_{\textrm{test}}(\tau)\bar{r}(\tau) = -\frac{k_{\tau}\bar{r}^2(\tau)}{V_{r}(\tau)} \sum_{\tau'}(\omega_{opt}(\tau',s) - 1)\pi_{\textrm{test}}(\tau')\bar{r}(\tau')
\end{align*}
By summing these expressions over $\tau$ , we reach:
\begin{align*}
\sum_{\tau'} \omega_{opt}(\tau',s)\pi_{\textrm{test}}(\tau')\bar{r}(\tau') = -\sum_{\tau}\frac{k_{\tau}\bar{r}(\tau)^2}{V_{r}(\tau)} \left(\lambda - \sum_{\tau'} \pi(\tau')\bar{r}(\tau')\right)
\end{align*}
and
\begin{align*}
\lambda = -\sum_{\tau}\frac{k_{\tau}\bar{r}^2(\tau)}{V_{r}(\tau)} \left(\lambda - \sum_{\tau'} \pi(\tau')\bar{r}(\tau')\right)
\end{align*}
Then,
\begin{align*}
\lambda = \frac{\sum_{\tau'} \frac{k_{\tau'}\bar{r}(\tau')^2}{V_{r}(\tau')}}{1 + \sum_{\tau'} \frac{k_{\tau'}\bar{r}(\tau')^2}{V_{r}(\tau')}} \sum_{\tau'} \pi_{\textrm{test}}(\tau')\bar{r}(\tau')
\end{align*}
Thus:
\begin{align*}
\omega_{opt}(\tau,s) = \frac{k_{\tau}\bar{r}(\tau)}{V_{r}(\tau)\pi_{\textrm{test}}(\tau)} \frac{\sum_{\tau'} \pi_{\textrm{test}}(\tau')\bar{r}(\tau')}{1 + \sum_{\tau'} \frac{k_{\tau'}\bar{r}(\tau')^2}{V_{r}(\tau')}}
\end{align*}
\end{proof}

\section{Proof: FIS dominates BIS}
The basic importance sampling (BIS) estimator can be written
\begin{align}
\widehat{J}_{\textrm{BIS}}(D^{N}) &= \frac{1}{N} \sum_{i=1}^N \int_{\tau} \frac{\pi_{\textrm{test}}(\tau)}{\pi_i(\tau)}r_i(\tau) 1_{\tau_i = \tau} \; ,
\end{align}
where $r_i(\tau)$ is drawn from a distribution with mean $\bar{r}(\tau)$ and variance $V_{r}(\tau)$. Similarly, the fused importance sampling (FIS) estimator can be written
\begin{align}
\widehat{J}_{\textrm{FIS}}(D^{N}) &= \sum_{i=1}^N \int_{\tau} \frac{\pi_{\textrm{test}}(\tau)}{\sum_j \pi_j(\tau)}r_i(\tau) 1_{\tau_i = \tau} \; .
\end{align}

We know that both estimators are unbiased so we focus on their variance. Using the law of total variance, we have that
\begin{align*}
\mathbb{V}[\widehat{J}_{\textrm{BIS}}] &= \mathbb{E}_R[\mathbb{V}_s[\widehat{J}_{\textrm{BIS}} | R]] + \mathbb{V}_R [\mathbb{E}_s[\widehat{J}_{\textrm{BIS}} | R]] \; .
\end{align*}
Let us start with the second term. Given the rewards, the expectation of the estimator is to be taken over the draws. We get:
\begin{align*}
\mathbb{E}[\widehat{J}_{\textrm{BIS}} | R] &= \frac{1}{N}\sum_{i=1}^N \int_{\tau} \frac{\pi_{\textrm{test}}(\tau)}{\pi_i(\tau)}r_i(\tau) \pi_i(\tau) \; d\tau = \frac{1}{N}\sum_{i=1}^N \int_{\tau} \pi_{\textrm{test}}(\tau) r_i(\tau) \; d\tau \; ,
\end{align*}
and the variance of this estimator is
\begin{align}
\mathbb{V}_R[\mathbb{E}[\widehat{J}_{\textrm{BIS}} | R]] &= \frac{1}{N} \int_{\tau} \pi_{\textrm{test}}^2(\tau) V_{r}(\tau) \; d\tau \; ,
\end{align}
Doing the same for FIS, we get
\begin{align*}
\mathbb{E}[\widehat{J}_{\textrm{FIS}} | R] &= \sum_{i=1}^N \int_{\tau} \frac{\pi_{\textrm{test}}(\tau)}{\sum_j \pi_j(\tau)}r_i(\tau) \pi_i(\tau) \; d\tau
\end{align*}
and the variance of this estimator is
\begin{align}
\mathbb{V}_R[\mathbb{E}[\widehat{J}_{\textrm{FIS}} | R]] &= \sum_{i=1}^N \int_{\tau} \frac{\pi_{\textrm{test}}^2(\tau)}{\left(\sum_j \pi_j(\tau)\right)^2}V_{r}(\tau) \pi^2_i(\tau) \; d\tau\nonumber\\
\mathbb{V}_R[\mathbb{E}[\widehat{J}_{\textrm{FIS}} | R]] &= \int_{\tau} \pi_{\textrm{test}}^2(\tau) V_{r}(\tau) \frac{\sum_i \pi^2_i(\tau)}{\left(\sum_j \pi_j(\tau)\right)^2}\; d\tau \; .
\end{align}

We now focus on the first term of the total variance. Since both BIS and FIS are averages over $i$, we compute the variance for each $i$ then average them.
\begin{align*}
\mathbb{V}[\widehat{J}_{\textrm{BIS}}^i | R] &= \int_{\tau} \frac{\pi_{\textrm{test}}^2(\tau)}{\pi_i^2(\tau)}r_i^2(\tau) \pi_i(\tau) \; d\tau - \left(\int_{\tau} \pi_{\textrm{test}}(\tau) r_i(\tau) \; d\tau\right)^2 \\
&= \int_{\tau} \frac{\pi_{\textrm{test}}^2(\tau)}{\pi_i(\tau)}r_i^2(\tau) \; d\tau - \left(\int_{\tau} \pi_{\textrm{test}}(\tau) r_i(\tau) \; d\tau\right)^2 \; .
\end{align*}
Thus, the variance of the global estimator is
\begin{align*}
\mathbb{V}[\widehat{J}_{\textrm{BIS}} | R] &= \frac{1}{N^2}\sum_i\int_{\tau} \frac{\pi_{\textrm{test}}^2(\tau)}{\pi_i(\tau)}r_i^2(\tau) \; d\tau - \frac{1}{N^2}\sum_i \left(\int_{\tau} \pi_{\textrm{test}}(\tau) r_i(\tau) \; d\tau\right)^2 \; .
\end{align*}
Taking the expectation over $r_i(\tau)$ yields
\begin{align*}
\mathbb{E}_R[\mathbb{V}[\widehat{J}_{\textrm{BIS}} | R]] &= \frac{1}{N^2}\int_{\tau} \pi_{\textrm{test}}^2(\tau)\left(\bar{r}^2(\tau) + V_{r}(\tau)\right)\sum_i \frac{1}{\pi_i(\tau)}\; d\tau
- \frac{1}{N}\left(\int_{\tau} \pi_{\textrm{test}}(\tau) \bar{r}(\tau) \; d\tau\right)^2 -\frac{1}{N}\int_{\tau} \pi_{\textrm{test}}^2(\tau) V_{r}(\tau) \; d\tau \; .
\end{align*}
Summing both terms for BIS, we get
\begin{align*}
\mathbb{V}[\widehat{J}_{\textrm{BIS}}] &= \frac{1}{N^2}\int_{\tau} \pi_{\textrm{test}}^2(\tau)\left(\bar{r}^2(\tau)+ V_{r}(\tau)\right)\sum_i \frac{1}{\pi_i(\tau)}\; d\tau - \frac{1}{N} \left(\int_{\tau} \pi_{\textrm{test}}(\tau) \bar{r}(\tau) \; d\tau\right)^2 \; .
\end{align*}
Let us know compute the conditional variance of FIS for one sample.
\begin{align*}
\mathbb{V}[\widehat{J}_{\textrm{FIS}}^i | R] &= \int_{\tau} \frac{\pi_{\textrm{test}}^2(\tau)}{\left(\sum_j \pi_j(\tau)\right)^2}r_i^2(\tau) \pi_i(\tau) \; d\tau - \left(\int_{\tau} \frac{\pi_{\textrm{test}}(\tau)}{\sum_j \pi_j(\tau)} r_i(\tau)\pi_i(\tau) \; d\tau\right)^2 \; .
\end{align*}
Taking the expectation over $R$ yields
\begin{align*}
\mathbb{E}_R[\mathbb{V}[\widehat{J}_{\textrm{FIS}}^i | R]] &= \int_{\tau} \frac{\pi_{\textrm{test}}^2(\tau)}{\left(\sum_j \pi_j(\tau)\right)^2}\left(\bar{r}^2(\tau) + V_{r}(\tau)\right) \pi_i(\tau) \; d\tau\\
 &\quad-\left(\int_{\tau} \frac{\pi_{\textrm{test}}(\tau)}{\sum_j \pi_j(\tau)} \bar{r}(\tau)\pi_i(\tau) \; d\tau\right)^2
 - \int_{\tau} \frac{\pi_{\textrm{test}}^2(\tau)}{\left(\sum_j \pi_j(\tau)\right)^2} V_{r}(\tau)\pi_i^2(\tau) \; d\tau \; .
\end{align*}
Summing over $i$ yields
\begin{align*}
\mathbb{E}_R[\mathbb{V}[\widehat{J}_{\textrm{FIS}} | R]] &= \int_{\tau} \frac{\pi_{\textrm{test}}^2(\tau)}{\sum_j \pi_j(\tau)}\left(\bar{r}^2(\tau) + V_{r}(\tau)\right) \; d\tau\\
 &\quad -\sum_i\left(\int_{\tau} \frac{\pi_{\textrm{test}}(\tau)}{\sum_j \pi_j(\tau)} \bar{r}(\tau)\pi_i(\tau) \; d\tau\right)^2 
 - \int_{\tau} \frac{\sum_i \pi_i^2(\tau)}{\left(\sum_j \pi_j(\tau)\right)^2}\pi_{\textrm{test}}^2(\tau) V_{r}(\tau) \; d\tau \; .
\end{align*}
Summing both terms for FIS, we get
\begin{align*}
\mathbb{V}[\widehat{J}_{\textrm{FIS}}] &= \int_{\tau} \frac{\pi_{\textrm{test}}^2(\tau)}{\sum_j \pi_j(\tau)}\left(\bar{r}^2(\tau) + V_{r}(\tau)\right) \; d\tau
-\sum_i\left(\int_{\tau} \frac{\pi_{\textrm{test}}(\tau)}{\sum_j \pi_j(\tau)} \bar{r}(\tau)\pi_i(\tau) \; d\tau\right)^2 \; .
\end{align*}

We may now compute the difference of the two variances:
\begin{align}
\mathbb{V}[\widehat{J}_{\textrm{BIS}}] - \mathbb{V}[\widehat{J}_{\textrm{FIS}}] 
&= \frac{1}{N^2}\int_{\tau} \pi_{\textrm{test}}^2(\tau)\left(\bar{r}(\tau) + V_{r}(\tau)\right)\sum_i \frac{1}{\pi_i(\tau)}\; d\tau 
 - \frac{1}{N}\sum_i \left(\int_{\tau} \pi_{\textrm{test}}(\tau) \bar{r}(\tau) \; d\tau\right)^2\nonumber\\
&\quad - \int_{\tau} \frac{\pi_{\textrm{test}}^2(\tau)}{\sum_j \pi_j(\tau)}\left(\bar{r}^2(\tau) + V_{r}(\tau)\right) \; d\tau \nonumber 
+\sum_i\left(\int_{\tau} \frac{\pi_{\textrm{test}}(\tau)}{\sum_j \pi_j(\tau)} \bar{r}^2(\tau)\pi_i(\tau) \; d\tau\right)^2\nonumber\\
&= \Bigg[\int_{\tau} \pi_{\textrm{test}}^2(\tau)\left(\bar{r}^2(\tau) + V_{r}(\tau)\right) 
 \left(\frac{1}{N^2}\sum_i \frac{1}{\pi_i(\tau)} - \frac{1}{\sum_j \pi_j(\tau)}\right)\; d\tau \Bigg]\nonumber\\
&\quad - \left(\frac{1}{N} \Bigg(\int_{\tau} \pi_{\textrm{test}}(\tau) \bar{r}(\tau) \; d\tau\right)^2 
 - \sum_i\left(\int_{\tau} \frac{\pi_{\textrm{test}}(\tau)}{\sum_j \pi_j(\tau)} \bar{r}(\tau)\pi_i(\tau) \; d\tau\right)^2\Bigg) \;.\nonumber
\label{eq:diff_variances}
\end{align}

We prove the positivity of the first term through a lemma:
\begin{lemma}
Let $a_1, \ldots, a_N$ $N$ strictly positive numbers. Then
\begin{align}
\frac{1}{N^2} \sum_i \frac{1}{a_i} &\geq \frac{1}{\sum_i a_i} \; .
\end{align}
\end{lemma}
\begin{proof}
Since both sides of the equation are strictly positive, we instead prove that the ratio of the two quantities is greater than 1.
\begin{align*}
\sum_i \frac{1}{a_i} \sum_j a_j &= \frac{1}{N^2} \sum_i\sum_j \frac{a_j}{a_i}
= \sum_i \frac{a_i}{a_i} + \sum_i \sum_{j>i} \left(\frac{a_j}{a_i} + \frac{a_i}{a_j}\right)
= N + \sum_i \sum_{j>i} \left(\frac{a_j}{a_i} + \frac{a_i}{a_j}\right)\\
&\geq N + \sum_i \sum_{j>i} 2
= N + 2\frac{N(N-1)}{2}
= N^2 \; .
\end{align*}
This concludes the proof.
\end{proof}
Using $a_i = \pi_i(\tau)$ and the positivity of $\pi_{\textrm{test}}^2(\tau)\left(\bar{r}^2(\tau) + V_{r}(\tau)\right)$, this proves the positivity of the first term.

To prove the negativity of the second term, we define a new random variable
\begin{align*}
z_i &= \int_{\tau} \frac{\pi_{\textrm{test}}(\tau)}{\sum_j \pi_j(\tau)} \bar{r}(\tau)\pi_i(\tau) \; d\tau
\end{align*}
where $i$ is taken uniformly at random in $[1, N]$. The variance of $z$ is:
\begin{align*}
V[z] &= \frac{1}{N} \sum_i \left(\int_{\tau} \frac{\pi_{\textrm{test}}(\tau)}{\sum_j \pi_j(\tau)} \bar{r}(\tau)\pi_i(\tau) \; d\tau\right)^2
- \left(\frac{1}{N} \sum_i \int_{\tau} \frac{\pi_{\textrm{test}}(\tau)}{\sum_j \pi_j(\tau)} \bar{r}(\tau)\pi_i(\tau) \; d\tau\right)^2\\
&= \frac{1}{N} \sum_i \left(\int_{\tau} \frac{\pi_{\textrm{test}}(\tau)}{\sum_j \pi_j(\tau)} \bar{r}(\tau)\pi_i(\tau) \; d\tau\right)^2
- \frac{1}{N^2}\left(\int_{\tau} \pi_{\textrm{test}}(\tau)\bar{r}(\tau)\; d\tau\right)^2 \; .
\end{align*}
Since $V[z]$ is positive, the second term of Eq.~\ref{eq:diff_variances} is negative.
Thus, $V[\widehat{J}_{\textrm{BIS}}] - V[\widehat{J}_{\textrm{FIS}}]$ is positive and FIS dominates BIS.

\section{Proof: Biasedness of EA when sampling policies depend on previous observed data}

We provide a small example why EA can be biased when the sampling policies depend on previous observed data. 

We consider a setting with one action whose reward is a Bernoulli of parameter 1/2. The policy stops sampling this action as soon as it experiences a 0.

We can compute analytically the expectation of the empirical average estimator. 

\begin{align*} 
\mathbb{E}(\hat{r}) = \sum_k (\frac{1}{2})^{k+1}\frac{k}{k+1} &= \sum_{k} (\frac{1}{2})^{k+1} - \sum_{k} \frac{1}{k+1}(\frac{1}{2}))^{k+1} = 1 - \ln(2)
\end{align*}
Thus, we underestimate the true reward of the action by ln(2) - 1/2.

\end{document}